\documentclass{article}

\usepackage{PRIMEarxiv}

\usepackage[utf8]{inputenc} 
\usepackage[T1]{fontenc}    
\usepackage{hyperref}       
\usepackage{url}            
\usepackage{booktabs}       
\usepackage{amsfonts}       
\usepackage{nicefrac}       
\usepackage{microtype}      
\usepackage{lipsum}
\usepackage{fancyhdr}       
\usepackage{graphicx}       
\graphicspath{{media/}}     

\usepackage{array}
\usepackage{caption}
\usepackage{subcaption}
\usepackage{sidecap}

\usepackage{authblk}
\usepackage{blindtext}

\title{VisBuddy - A Smart Wearable Assistant for the Visually Challenged

}

\author[1]{Ishwarya Sivakumar }
\author[1]{Nishaali Meenakshisundaram}
\author[1]{Ishwarya Ramesh }
\author[1]{\\Shiloah Elizabeth D}
\author[2]{Sunil Retmin Raj C}
\affil[1]{Department of Computer Science and Engineering, CEG Campus, Anna University.}
\affil[2]{Department of Information Technology, MIT Campus, Anna University.}
\affil[*]{\textit{Email: shiloah@annauniv.edu}}

\begin{document}
\maketitle

\begin{abstract}
Vision plays a crucial role in comprehending the world around us. More than 85\% of the external information is obtained through the vision system. It influences our mobility, cognition, information access, and interaction with the environment and other people. Blindness prevents a person from gaining knowledge of the surrounding environment and makes unassisted navigation, object recognition, obstacle avoidance, and reading tasks significant challenges. 
Many existing systems are often limited by cost and complexity. To help the visually challenged overcome these difficulties faced in everyday life, we propose VisBuddy, a smart assistant to help the visually challenged with their day-to-day activities. VisBuddy is a voice-based assistant where the user can give voice commands to perform specific tasks. It uses the techniques of image captioning for describing the user's surroundings, optical character recognition (OCR) for reading the text in the user's view, object detection to search and find the objects in a room and web scraping to give the user the latest news. VisBuddy has been built by combining the concepts from Deep Learning and the Internet of Things. Thus, VisBuddy serves as a cost-efficient, powerful, all-in-one assistant for the visually challenged by helping them with their day-to-day activities. 

\end{abstract}

\keywords{visual assistant \and image captioning \and optical character recognition \and object detection \and raspberry pi \and wearable system}

\section{Introduction}
Visual impairment or blindness is one of the common disabilities worldwide. The leading causes of blindness are age-related diseases, accidents and congenital disorders. According to the World Health Organization, around 2.2 billion people worldwide suffer from near or distance vision impairment. Blindness is the lack of visual perception due to neurological or physiological factors. Blindness poses many challenges in day-to-day activities like unassisted navigation, object recognition, obstacle avoidance, and reading tasks. It also prevents the person from gaining knowledge about the surrounding environment. As a result, they cannot see and feel the world like others.

Currently, white cane and guide dogs are used by the visually challenged for assistance in mobility. The white cane allows the holder to identify the obstacles and orientation marks by scanning the surroundings. The white cane with a red tip is considered the international symbol of blindness. It is simply a long cane used to extend the user's range of touch sensation. There are different types of white canes with specific purposes like long cane, guide cane, identification cane, kiddie cane, support cane etc. The guide dogs are trained to guide a blind person by navigating around various obstacles and indicating when necessary to go up or down a step. However, only a few people employ guide dogs because of the financial expenses involved in taking care of the dog. It also takes time to match a guide dog to a visually challenged person. Moreover, since these dogs are red-green color blind, they cannot decipher the traffic signs. Visually challenged people can also use Human echolocation for navigation. It is the practice in which the person actively creates sounds and detects the objects in the surroundings by sensing the echoes from those objects.

The recent technological advancements have facilitated the creation of newer ways and devices to improve the lives of the visually challenged. For example, GPS devices can also be used as mobility aids. Recent advances in the Internet of Things(IoT) has increased the power of edge computing devices. Latest Deep Learning algorithms have increased the efficiency of image processing and natural language processing tasks. In addition, cloud Computing has augmented the computing power available to individuals. These growing technologies can help create better and efficient assistants for the visually challenged and help them make their lives easier. 

  This paper proposes an assistive device that can perform multiple information cognition tasks at once and is easy to use, affordable, and lightweight. The assistant is controlled by voice commands from the user. Combining concepts from Deep Learning and the Internet Of Things, we have developed highly optimized scene recognition, object detection, optical character recognition and online newspaper reading modules. These modules are used to process the information about the user's surroundings and convey them to the user. 

\section{Related Works}
Development of assistive technology for the visually challenged has gained traction in the recent years. This technology is mainly focused on helping the user with assistance for navigation, information about the surroundings and accessing digital devices. The first class of devices come under mobility assistance and scan the user's surrounding and convey the information to the user to help in navigation and obstacle avoidance. The second class of devices use technologies like Optical Character Recognition (OCR) to process the information and inform the user. The third class of devices include Voice Synthesizers, Braille Output terminals and magnifiers that help the user to read and know the information.
\\
A smart wearable glass has been developed using Raspberry Pi (RPi) and Ultrasonic sensor by Khan \emph{et al} \cite{r1}. In their work, they have developed modules for obstacle detection and reading assistant. In the works of Hengle \emph{et al} \cite{r2}, a smart cap has been developed using RPi. The smart cap can be used to describe the user's surroundings using image captioning, recognize faces using dlib face recognition, read text using Google Vision API and fetch news from the internet. 
\\
With the emergence of smartphones and the mobile operating systems like Android and iOS, smartphone based features have been developed for helping the visually challenged. A live location tracker app has been presented by Khairnar \emph{et al}\cite{r3}. Manduchi \emph{et al}\cite{r9} has experimented using the mobile camera for sign based wayfinding to detect specific color markers. An android mobile app has been developed using Google Cloud Services for image recognition, currency recognition, object recognition by Kavya \emph{et al}\cite{r10}
\\
Using multiple sensors to avoid obstacles and help in navigation is also popular. These sensors are worn by the user or placed on another object used by the person like a white cane. The NavGuide\cite{r11} uses multiple sensors to detect obstacles upto the knee level. The wearable device proposed by Katzschmann \emph{et al}\cite{r12}  uses a sensor belt and haptic strap. The sensor belt is an array of time-of-flight sensors and use pulses of infrared light to provide reliable measurements of the distance between the user and surroundings. The haptic strap provides haptic feedback and helps the user to navigate. The NavCane\cite{r13} uses several sensors placed on a white cane to help the user navigate the surroundings. It helps the user to identify the obstacle free paths in both indoor and outdoor settings.  
\\
Previous works have problems related to affordability and ease of use. In addition, these works have not implemented multiple essential functionalities in the same device. We propose a system that incorporates four essential functionalities that aid in information processing on the same device. This system is low-cost, lightweight, easy-to-use and efficient in its functioning. The novel features that have been implemented in this paper are 
\begin{enumerate}
    \item An Object Discovery Module using object detection algorithms and novel heuristics.
    \item Dynamic column determination algorithm to detect any number of columns of text in the image given to the reading assistant. Previous works could identify up to 2 columns only.
    \item Use of the Inception-V3 model as the encoder for the scene recognition module to get richer feature representations. Previous works have used VGG16 and ResNet150. 
\end{enumerate}

Though the current prototype of the system does not have advanced features like navigation and detection of obstacles, the amenable design allows for future enhancements. 

\section{Proposed System}
VisBuddy is a wearable assistant that takes in voice commands and images as input and gives an audio output to the user. The wearable assistant has been built using Raspberry Pi (RPi) 4 Model B with a 2 GB RAM and 16 GB ROM (using micro-SD card), Raspberry Pi Camera (8MP), Power Bank (3.7V, 10000 mAH), USB Microphone and USB Earphones. The 10000maH energy source lasts up to 12 hours on a single charge. The Raspberry Pi has been placed in a case for protection and uses heat sinks and a fan to regulate the heating.  
The RPi case and power bank are placed in a hip pouch that the user can wear. The camera is attached to the outside of the hip pouch. The earphones and microphone are attached to the user. Placing the RPi in a hip pouch protects the user from harmful radiation emitted by RPi. It also provides the camera with a clear field of view and is comfortable to wear. 
\\
In Fig 1, the overall functioning of the device is shown. The main programming logic and files are stored on the RPi. The modules are triggered by voice commands from the user. The speech commands are converted to text using the Python Speech Recognition library powered by Google Speech Recognition. \\
\label{sysdesign}


\begin{figure*}[!b]
\centering{\includegraphics[width=16cm,height=14cm]{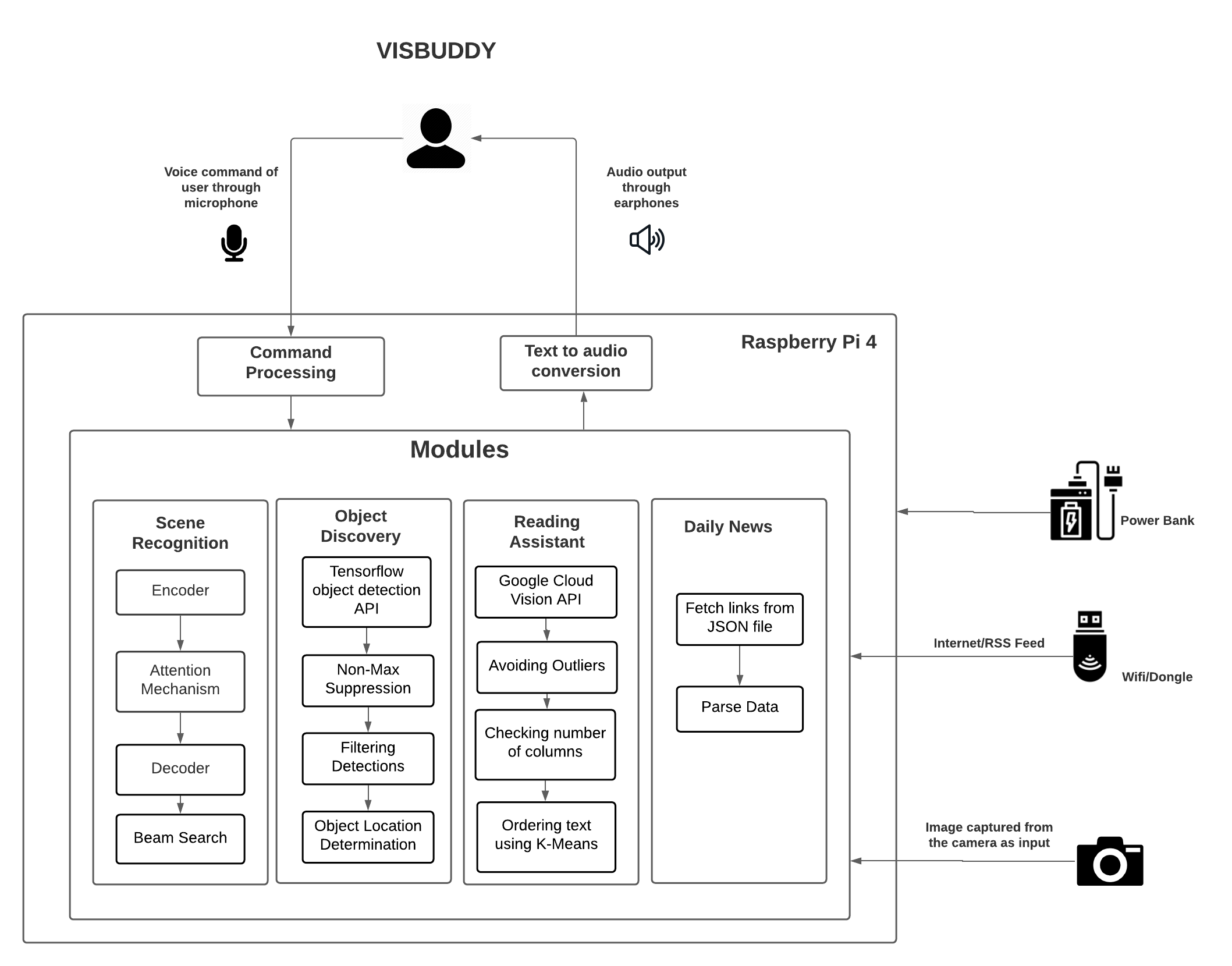}}
\caption{Full System Architecture}
\end{figure*}

The command is then parsed to identify the module to be run. The Scene Recognition, Object Discovery and Reading Assistant modules require an input image taken using the onboard RPi camera. The script corresponding to the recognized module is invoked and executed. If the command given by the user is invalid or not recognized, then the user is instructed to give the command again. The output text from the modules is converted to audio and conveyed to the user through the earphones using the eSpeak library. All the outputs are conveyed to the user in English. 

\section{System Design and Implementation}
\subsection{Scene Recognition}
This module generates a textual description of what is in front of the user. It is triggered by voice commands like \emph{"what is in front of me"} from the user. First, the camera then takes a picture of the user's view. Then, image is processed by the scene recognition model proposed in this paper that is based on the attention-based encoder-decoder architecture proposed by Kelvin Xu et al \cite{r4} in their paper "Show, Attend and Tell". Finally, the caption is converted into an audio output and conveyed to the user.

\subsubsection{Encoder}
The encoder is a Convolutional Neural Network (CNN) that generates a fixed-size feature vector for the input image. Instead of the VGG16 model\cite{r17} used by \cite{r4} or the ResNet-101 model used by \cite{r1} or the MobileNet-V2\cite{r18} that has been designed for edge devices, this paper proposes the use of Inception-V3 model\cite{r8} trained on the Imagenet dataset as the encoder. This has been chosen because it acts as a multiple feature extractor by computing 1x1, 3x3 and 5x5 convolutions within the same module of the inception network and gives a richer feature representation of the image. Furthermore, it has a lesser computational cost than ResNet-101. Since the image has to be encoded and not classified, the last two layers, the pooling and the linear layer, have been discarded. The comparative performance of Inception-V3 against the other models is discussed in the Results and Analysis section.
\\The RGB image from the camera is resized to 299x299 pixels and passed to the encoder. It produces an 8x8x2048 feature vector which is reshaped to 64x2048. Each pixel represents a part of the original image and the information is encoded in the 2048 dimensional vector. This feature vector is given to the decoder to generate the caption.

\subsubsection{Decoder with Visual Attention}
The decoder generates the caption for the image based on the feature vector from the encoder. The decoder model proposed in this paper has been built using Long Short Term Memory Cell (LSTM)\cite{r6} with Soft Attention mechanism. The LSTM was chosen instead of Gated Recurrent Unit (GRU)\cite{r15} because it was more adept at handling dependencies across longer sentences and the vanishing and exploding gradients problem. When Attention Mechanism is not used, the decoder looks at the entire image for generating each word of the caption. However, this approach is inefficient as each word is influenced only by a specific part of the image. Hence by using an attention mechanism, the decoder considers only the particular region of the image that influences the word and the previous word to generate the word at the current timestep. This gives a more accurate caption for the image. The context vector denotes the importance of each part of the image for that timestep. Soft Attention has been used as it is more deterministic than Hard Attention, and the pixels' weights add up to 1. 
\\
The Teacher Forcing strategy has been used to aid the training process. The encoder representation of the image and the internal states of the LSTM from the previous time step are used to calculate the context vector.  Then the LSTM decoder takes the context vector and the original word at the previous timestep to generate the next word. This process is continued until the decoder reaches the end of the caption or a pre-specified caption length.

\subsubsection{Beam Search}
Beam Search has been used for caption generation since the Greedy Search does not yield optimal results. In Greedy Search, the entire sequence depends on the first word that has been generated. Whereas, in Beam Search, at each time step, k most likely sequences are kept track of and finally, the sequence with the highest score is chosen. It was found that the best performance was achieved with a beam width k=3 for the model developed in this paper.

\subsubsection{Model Training}
The image captioning encoder-decoder model has been trained on the Flickr8k dataset. The concept of Transfer Learning has been used for the encoder. The weights of the Inception-V3 model trained on the Imagenet dataset were used. The RGB image taken by the camera is first resized to 299x299 pixels. Then it is preprocessed using the default preprocessing function for Inception-V3 model available in the tf.keras.applications.inception\_v3 package. It is then given to the encoder to generate the feature vector for the image. The models were developed using Tensorflow and trained in Google Colab.

\subsection{Object Discovery}
This module is used to assist the user in finding an object in the surrounding. The module is activated by voice commands like "\emph{Where is my cellphone}" from the user. In this command, the cellphone is the object to be searched. First, the camera takes a picture of what is in front of the user. Second, this image is then passed to the Object Detection Model, implemented using Tensorflow Object Detection API. Finally, the model's outputs containing the bounding box coordinates of the different objects in the image are processed using novel heuristics to find the object and its location. The location of objects in the image is classified into three regions: left side, right side and straight view. 

\subsubsection{Object Detection}
The RetinaNet model \cite{r5} and Single Shot Detector (SSD) with MobileNet\cite{r16} were considered because of their low inference time, high accuracy and small size compared to other models. The performance of these two object detection models on real-time images was compared to choose the better-performing model for implementation. The model weights were loaded from the Tensorflow Detection Model Zoo. After comparing the results obtained from both the models, it was observed that RetinaNet had better performance on the real-time test images. Hence, RetinaNet was chosen for the implementation of the module.

\subsubsection{Finding and Locating the object}
The object's name to be searched is extracted from the user command. The outputs from the object detection model are filtered using the index of the class label of the object and a confidence score of 0.35 as a threshold. The object's location (left, right or straight) is obtained by computing the x-coordinate of the center of the bounding box. This has been used because it gives a near accurate position of the object. The exceptional cases like the absence of the requested object in the image or the requested object not being present in the available classes of objects have also been handled. 

\subsection{Reading Assistant}
This module reads the text detected in the user's view. It is activated by voice commands like \emph{"Can you read the text?"} from the user. It has been built using Google Cloud Vision API. First, the camera takes a picture of the user's view. Second, the image is processed using the Vision API. This returns the detected text structured into pages, blocks, paragraphs, words and symbols, along with their corresponding x and y coordinates. The outliers in the detected text are removed since they are not required. Since the recognized text is not in the correct top to bottom, left to right order, a novel ordering algorithm has been used to order it. This algorithm uses Z-Score and K-means clustering algorithms. The performance of Vision API for this task was better than Tesseract which was used in \cite{r14}. Hence it was chosen for the implementation of this module.

\subsubsection{Avoiding outliers using Z-score}
Z-score describes the position of the text blocks from the mean position of all the text blocks in the image. It is computed using the x-coordinate of the midpoint of each text block.  If the absolute value of the Z-score of a block is greater than or equal to the threshold of 1.75, then it is considered an outlier and removed. This process is repeated for the remaining text blocks until all the outliers have been removed. The Z-score threshold has been chosen as a heuristic value.  If there is only one single-column text region, the text is read out to the user. Else, the text is further processed. 

\subsubsection{K-means Clustering for Ordering Multi-Columnar Text}
The K-means clustering algorithm has been used to order texts with more than one column. This idea was inspired by the works of Hengle \emph{et al}\cite{r2}. The output from the Vision API returns blocks of detected text. First, Auto clustering is done to find the number of columns of text. This is done by calculating the distance between two consecutive blocks of text. If this distance is greater than the threshold of 150, then the blocks are designated as different columns. The distance threshold has been chosen as a heuristic value. Then, the number of columns has been used as the cluster size for the K-means clustering algorithm. K-means clustering orders the text into columns. It also arranges the blocks from left to the right. The text blocks in different columns are then ordered from top to bottom by sorting their coordinates. Thus, this process enables the assistant to recognize images with multiple columns of text also.

\subsection{Online Newspaper Reader}
The Daily News module aims to provide the current news to the user. The  Newspaper3k and Feedparser libraries have been used for news scraping. However, the commonly used BeautifulSoup library has not been considered because of its latency issues.
\\
Voice commands from the user such as \emph{"What is in the news?"} activates the module. A JSON file with popular newspaper names, RSS feeds, and URLs is stored on the RPi. First, the module checks if an RSS feed for a particular source is available. If yes, the news articles are downloaded using the Feedparser library. The articles that have the publishing date are only considered to get the latest articles. 
\\
If the RSS feed is not available, then the news articles are downloaded using the website URLs using the Newspaper3k library. A maximum of 5 articles is downloaded for each domain. The headlines of the downloaded news articles are parsed and stored in a text file which is then read to the user.

\section{Results and Analysis}
\subsection{Scene Recognition}
The scene recognition module is assessed by evaluating the coherence and semantic meaning of the captions. Towards this purpose, BLEU (Bilingual Evaluation Understudy) Score\cite{r7} has been used. The Attention-based Encoder-Decoder model has been evaluated using BLEU-1, BLEU-2, BLEU-3 and BLEU-4. The comparison of the performances between ResNet-101, MobileNet-V2 and Inception-V3 has been tabulated in Table-1. Some sample captions generated by the Scene Recognition model with Inception-V3 has been shown in Fig 2.

\begin{table}[hbt]
\caption{Scene Recognition Metrics}
\centering
\begin{tabular}{p{0.18\linewidth} c c c c }
\hline\noalign{\smallskip}
Model & BLEU-1 & BLEU-2 & BLEU-3 & BLEU-4\\
\noalign{\smallskip}\hline\noalign{\smallskip}
Inception-V3 encoder & 0.63 & 0.41 & 0.27 & 0.17\\
\noalign{\smallskip}\hline\noalign{\smallskip}
ResNet-101 encoder & 0.61 & 0.39 & 0.25 & 0.15\\
\noalign{\smallskip}\hline\noalign{\smallskip}
MobileNet-V2 encoder & 0.50 & 0.24 & 0.13 & 0.07 \\
\noalign{\smallskip}\hline\noalign{\smallskip}

\end{tabular}
\end{table}

The results clearly show that Inception-V3 produces the best captions even though it is bigger than Resnet-101 and MobileNet-V2. Hence has been chosen for the implementation of this module. The average latency to generate a caption is 3s.

\begin{figure}
     \centering
     \begin{subfigure}[b]{0.23\textwidth}
         \centering
         \includegraphics[width=\textwidth, height=5cm]{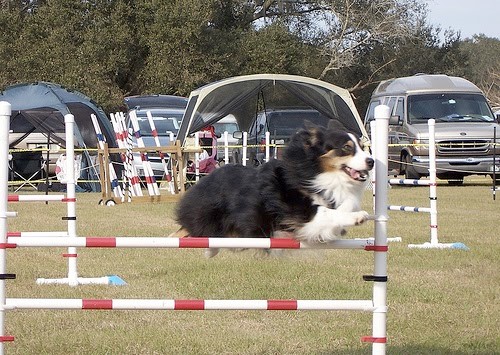}
         \caption{black dog is jumping over fence}
         \label{doggie}
     \end{subfigure}
     \hfill
     \begin{subfigure}[b]{0.23\textwidth}
         \centering
         \includegraphics[width=\textwidth, height=5cm]{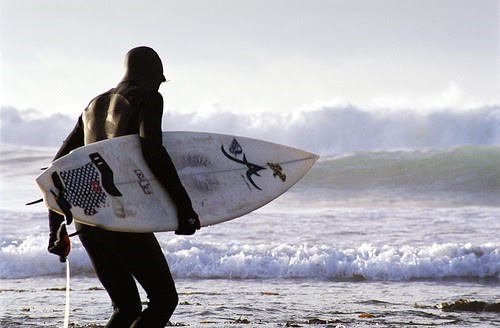}
         \caption{man in black wetsuit is standing in the beach}
         \label{diver}
     \end{subfigure}
     \hfill
     \begin{subfigure}[b]{0.23\textwidth}
         \centering
         \includegraphics[width=\textwidth, height=5cm]{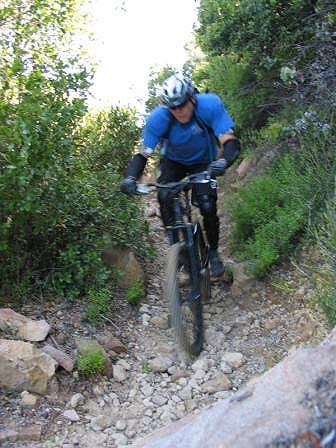}
         \caption{the man is riding the bike through the woods}
         \label{cyclist}
     \end{subfigure}
     \hfill
     \begin{subfigure}[b]{0.23\textwidth}
         \centering
         \includegraphics[width=\textwidth, height=5cm]{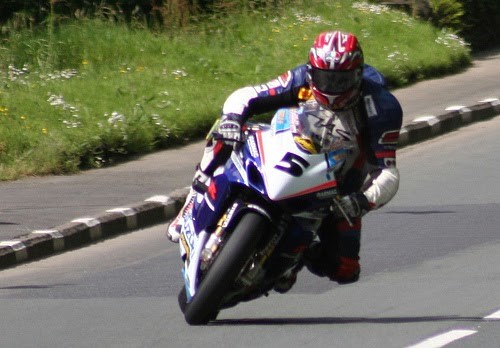}
         \caption{motorcycle racer is riding on motorcycle}
         \label{biker}
     \end{subfigure}
     
        \caption{Scene Recognition Outputs}
        \label{scene recog outputs}
\end{figure}

\subsection{Object Discovery}
The comparative performance of RetinaNet and MobileNet for the object discovery task has been tabulated in Table-2. It is evident from it that RetinaNet is the better performing model. Sample outputs of the Object discovery module are shown in Table-3.
\\Furthermore, Retinanet was faster when compared to MobileNet. Retinanet gave results in 46ms compared to 48ms for MobileNet. The module was implemented with the RetinaNet object detection model based on these results.
\\The module was tested in different real-time scenarios to analyze its performance. Table-3 shows the output of the module in different scenarios.

\begingroup
\setlength{\tabcolsep}{6pt} 
\renewcommand{\arraystretch}{1} 

\begin{table}[hbt!]
\caption{Object Discovery Models Comparison}
\centering

\begin{tabular}{m{.3\linewidth} m{.275\linewidth} m{.2\linewidth}}

\noalign{\smallskip}\hline\noalign{\smallskip}
\centering
Input & 
\begin{center}Output with RetinaNet \end{center}
 &
 \begin{center} Output with SSD \end{center} \\
\noalign{\smallskip}\hline\noalign{\smallskip}

\centering
\includegraphics[width=2.5cm, height=2cm]{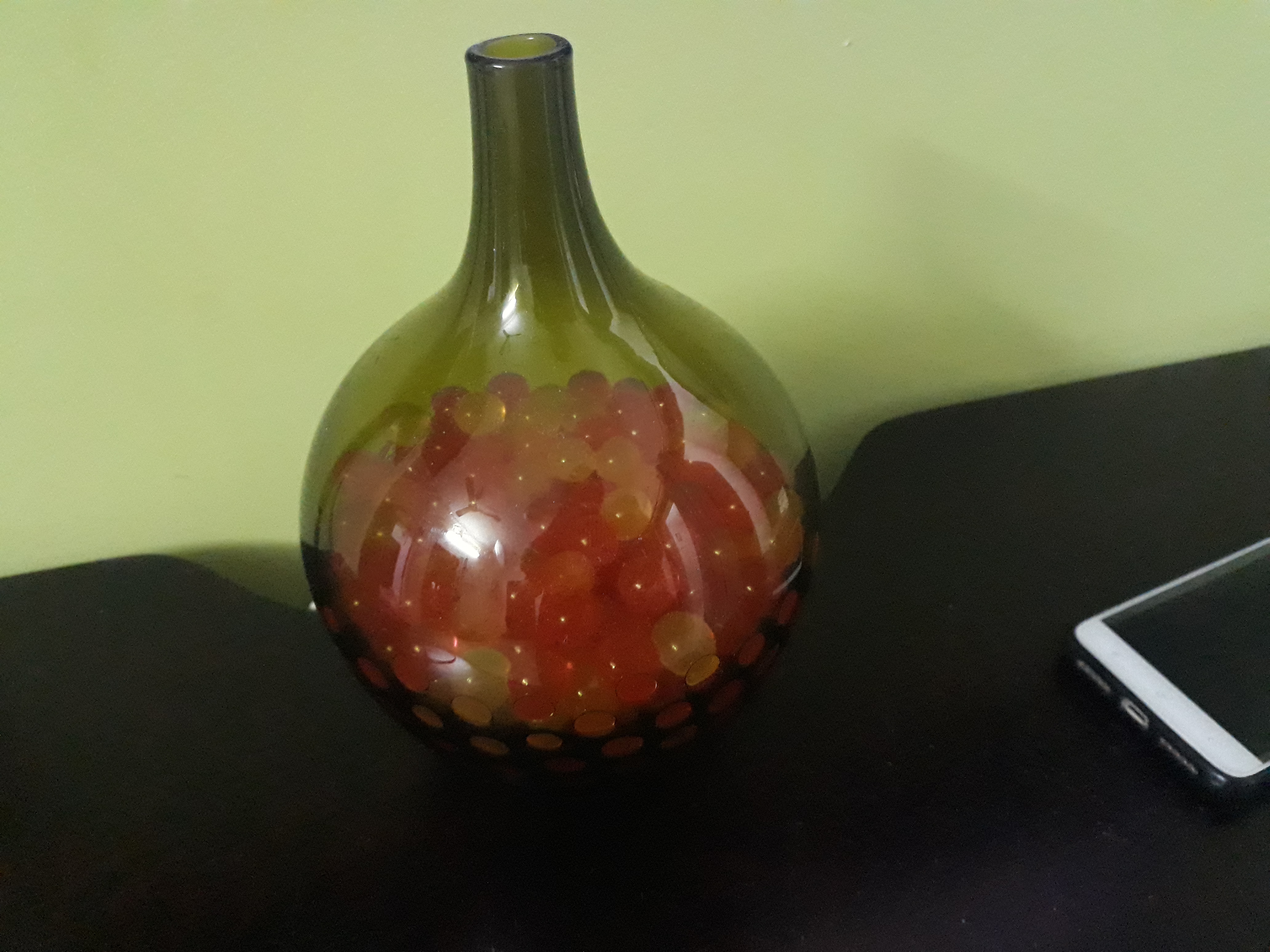}
 & 1 cell phone on your right & cell phone not found \\
  \noalign{\smallskip}\hline\noalign{\smallskip}

\centering
\includegraphics[width=2.5cm, height=2cm]{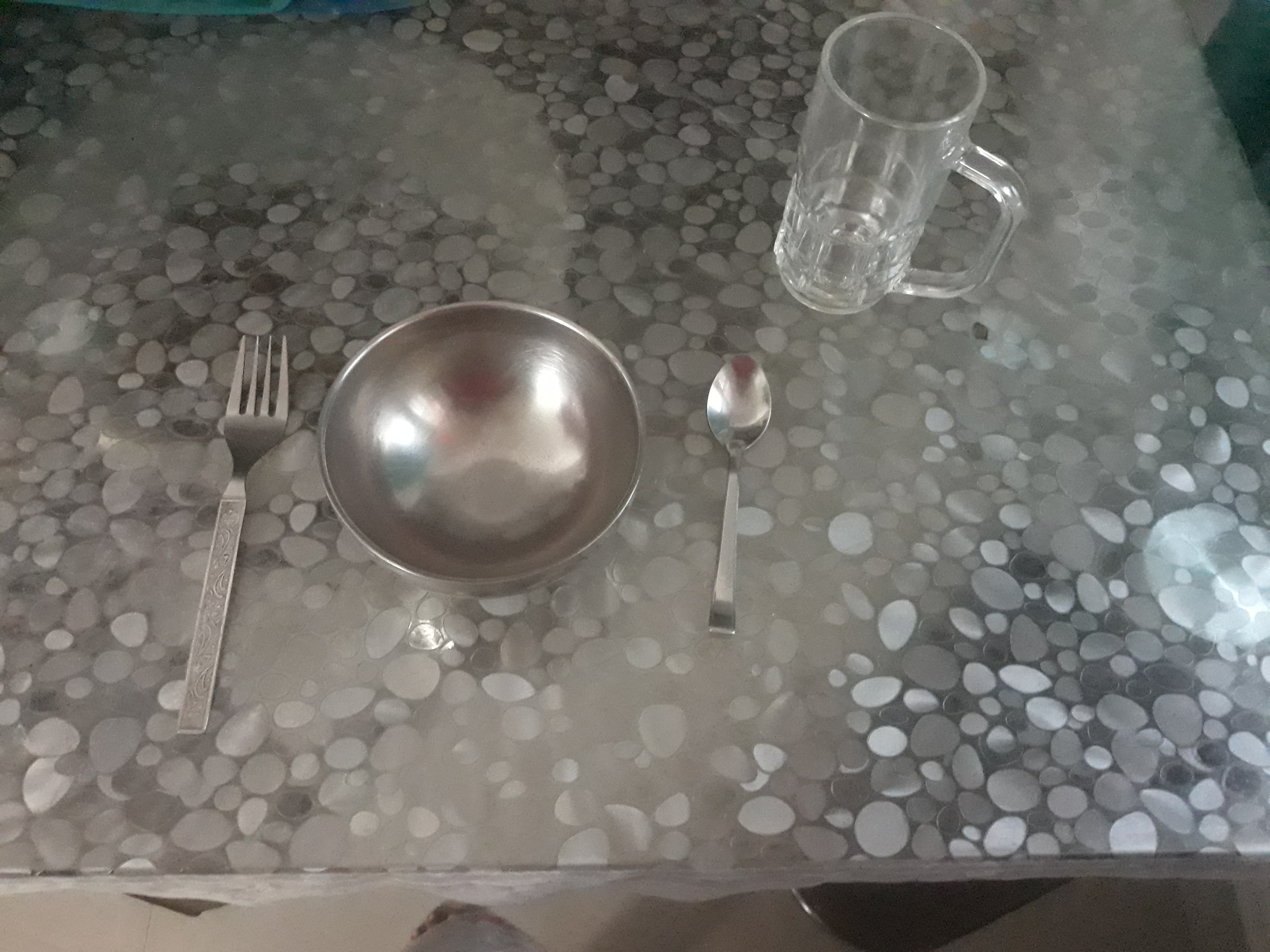}
 & 1 spoon on your straight & spoon not found \\
\noalign{\smallskip}\hline\noalign{\smallskip}

\centering
\includegraphics[width=2.5cm, height=2cm]{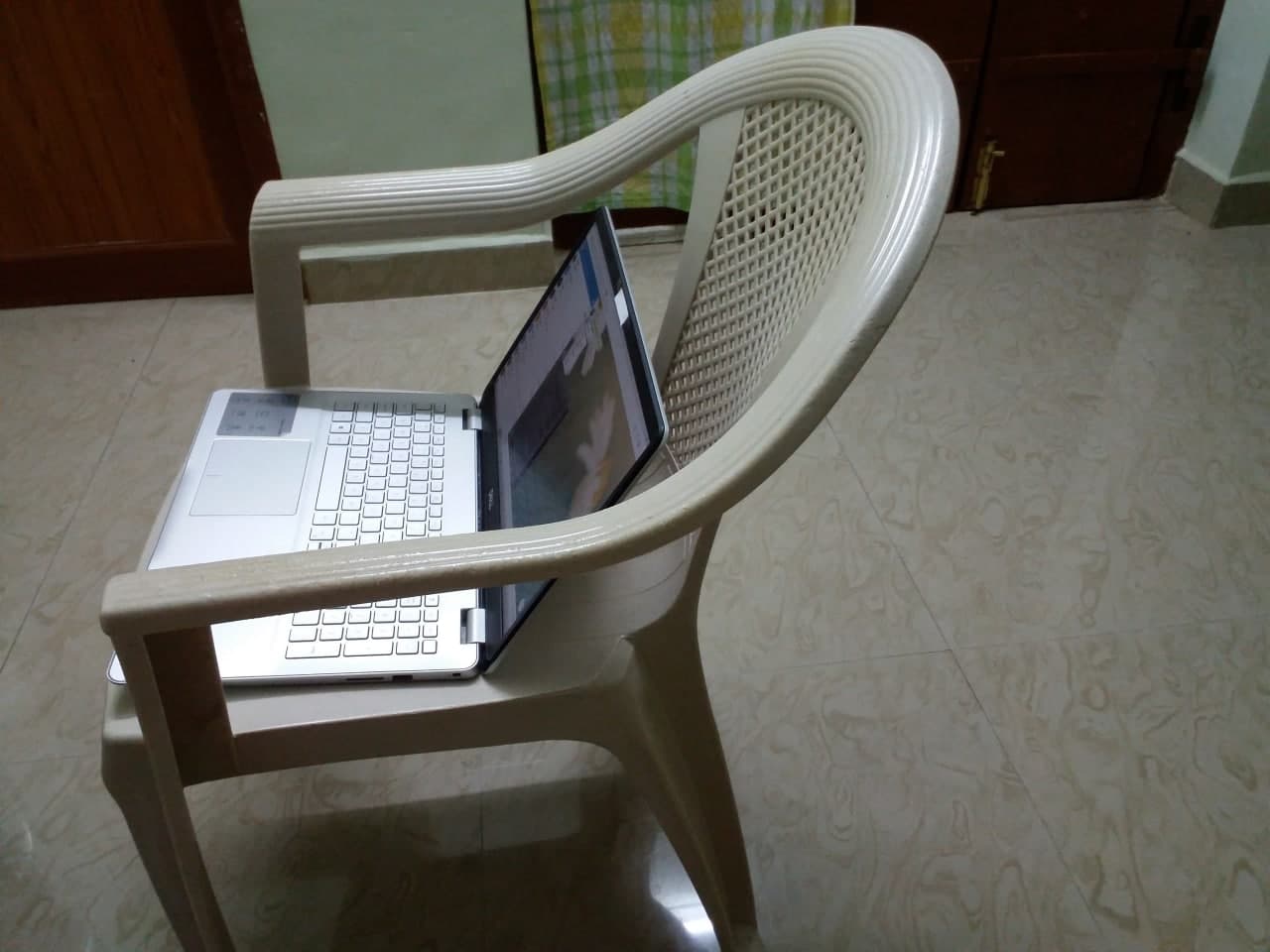}
 & 1 laptop on your straight & laptop not found \\
\noalign{\smallskip}\hline\noalign{\smallskip}

\centering
\includegraphics[width=2.5cm, height=2cm]{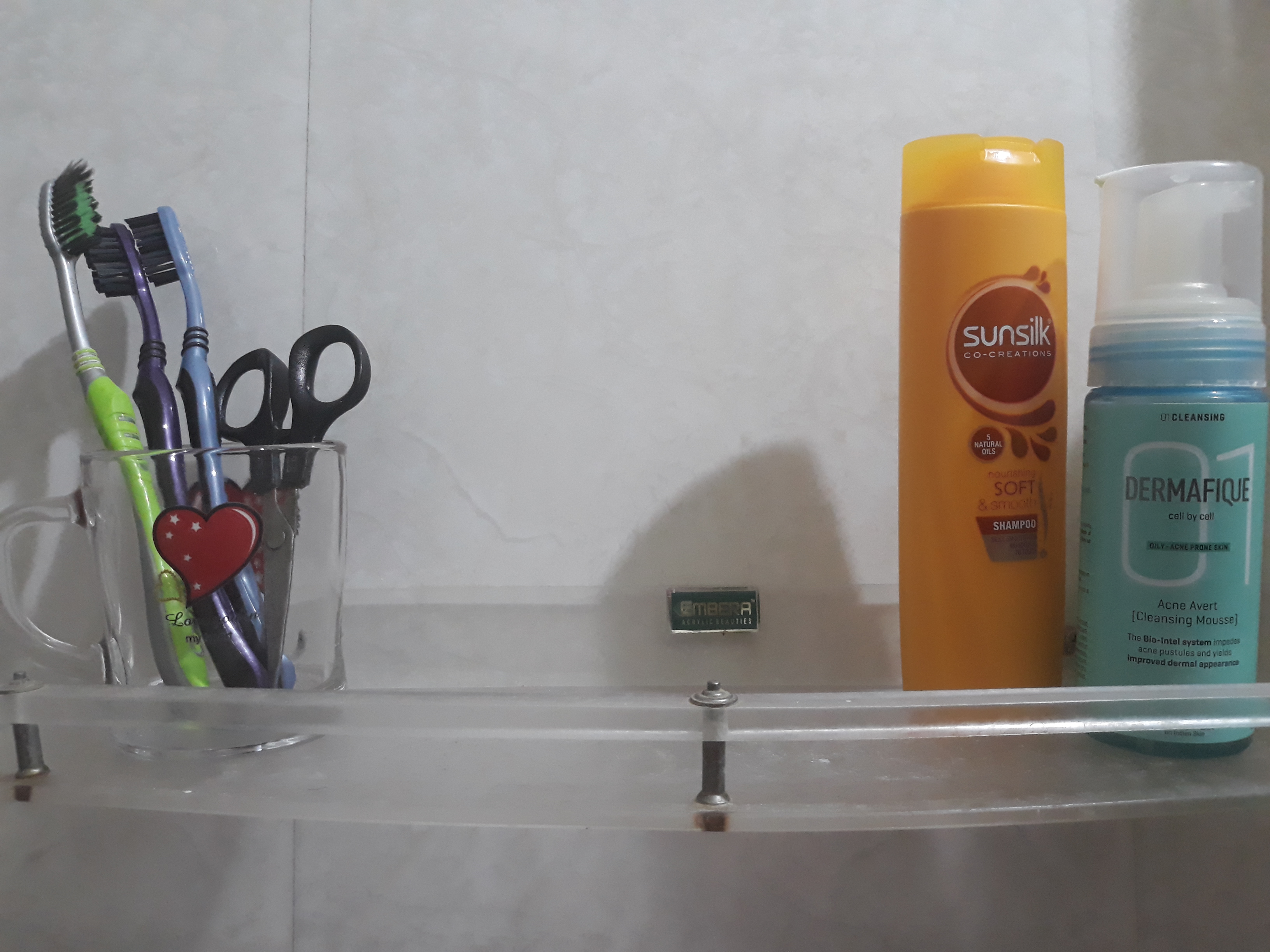}
 & 1 scissors on your left & scissors not found \\
\noalign{\smallskip}\hline

\end{tabular}
\end{table}

\endgroup

\begingroup
\setlength{\tabcolsep}{10pt} 
\renewcommand{\arraystretch}{1.5} 

\begin{table}[hbt!]
\caption{Sample Outputs Of The Object Discovery Module}
\centering
\begin{tabular}{m{.3\linewidth} m{.20\linewidth} m{.20\linewidth}}
\noalign{\smallskip}\hline\noalign{\smallskip}

\centering
Input image & Input command & Output \\
\noalign{\smallskip}\hline\noalign{\smallskip}

\centering
\includegraphics[width=2.5cm, height=2cm]{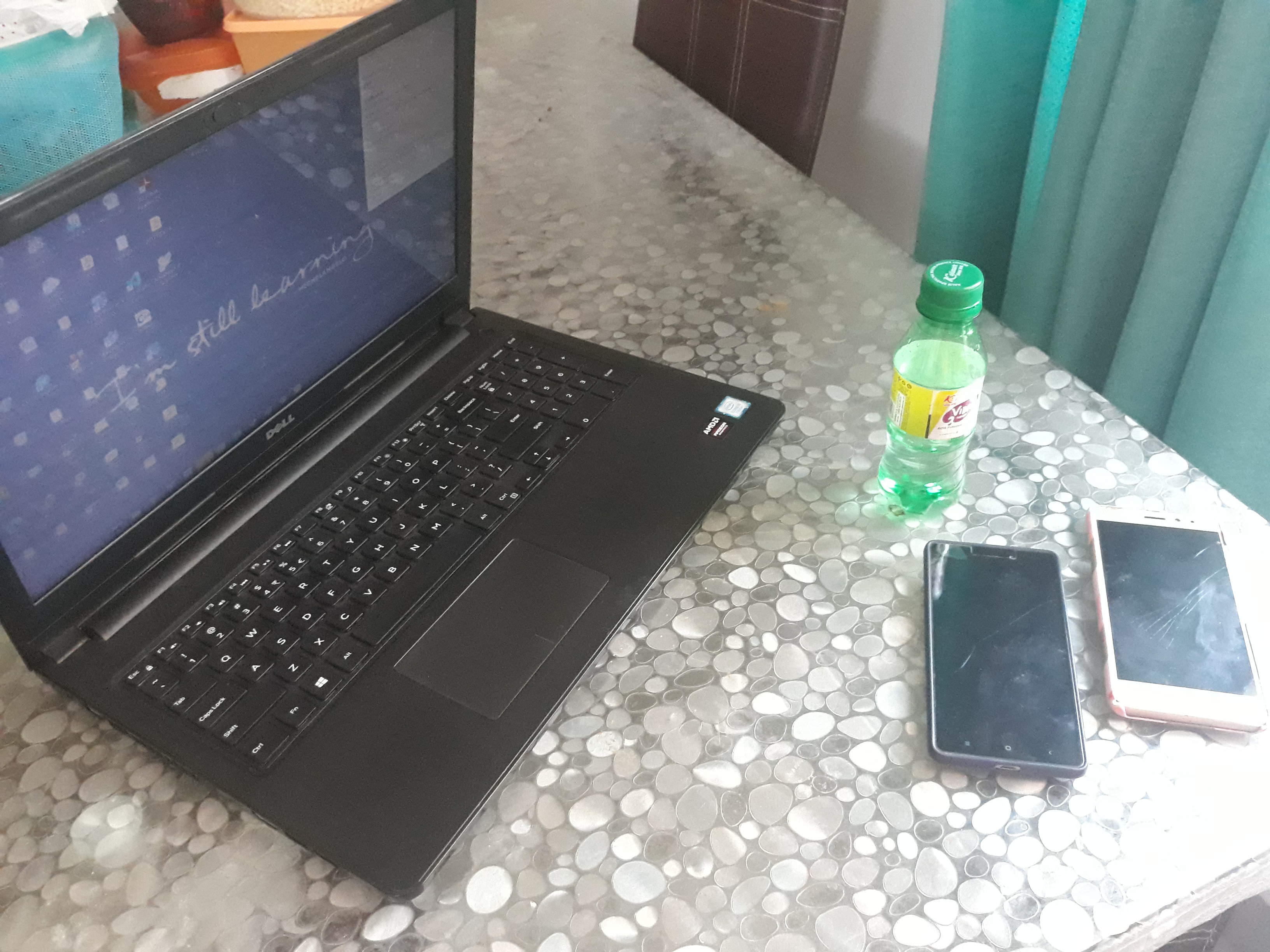}
 & can you find the laptop & 1 laptop on your left \\
\noalign{\smallskip}\hline\noalign{\smallskip}

\centering
\includegraphics[width=2.5cm, height=2cm]{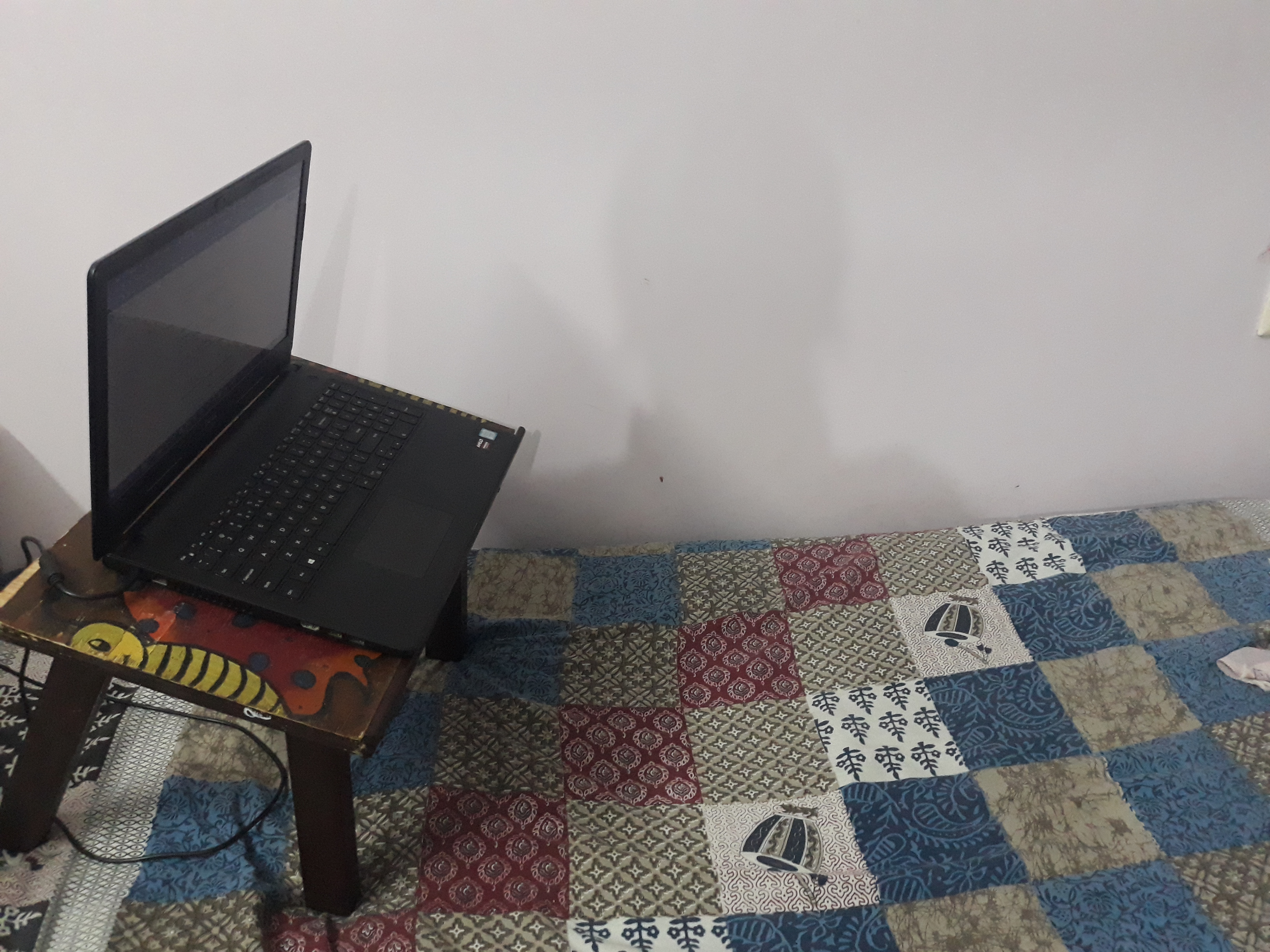}
 & can you find my laptop & 1 laptop on your left \\
\noalign{\smallskip}\hline\noalign{\smallskip}

\centering
\includegraphics[width=2.5cm, height=2cm]{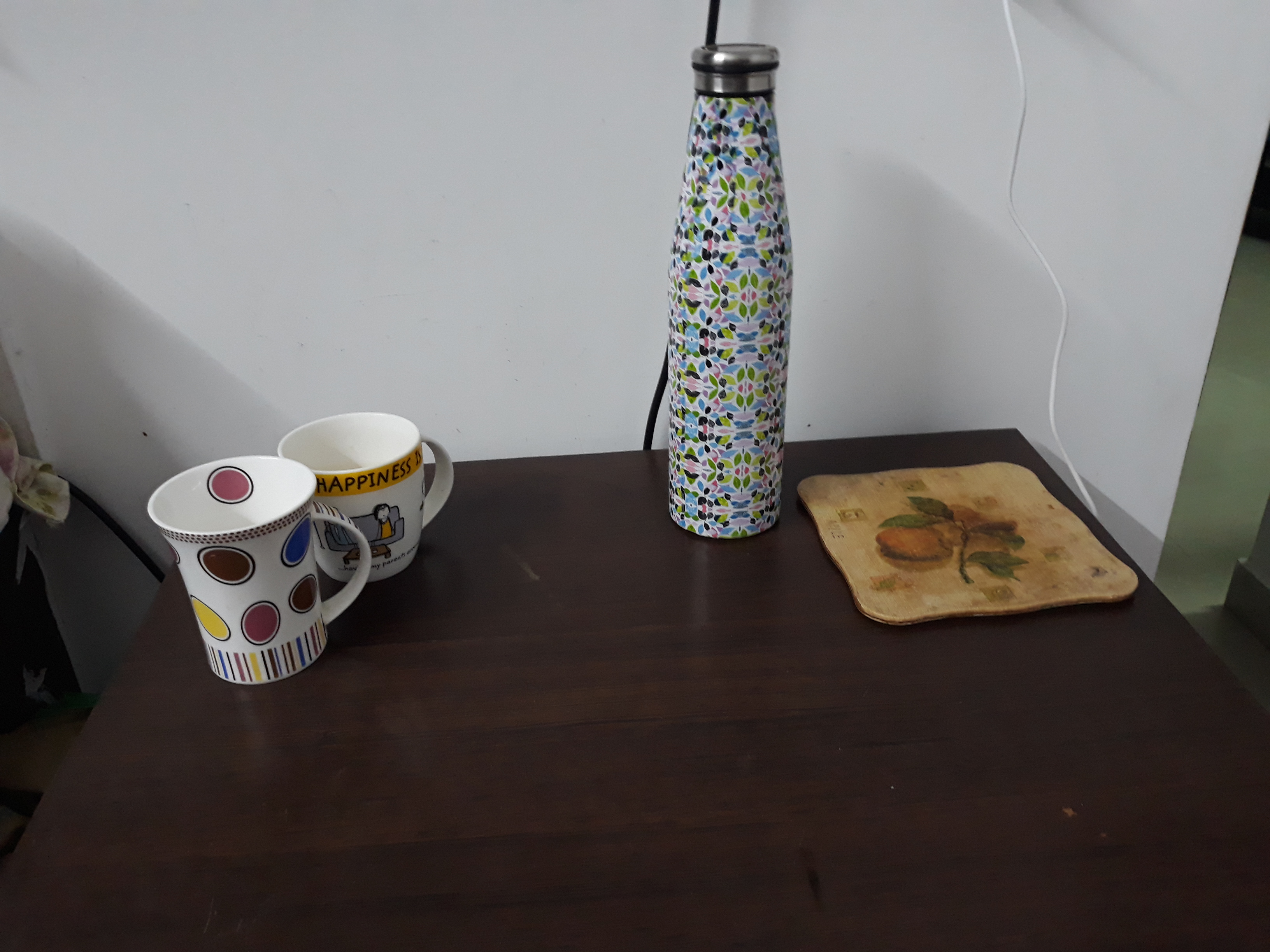}
 & can you find my cup & 2 cup on your left \\
\noalign{\smallskip}\hline\noalign{\smallskip}

\centering
\includegraphics[width=2.5cm, height=2cm]{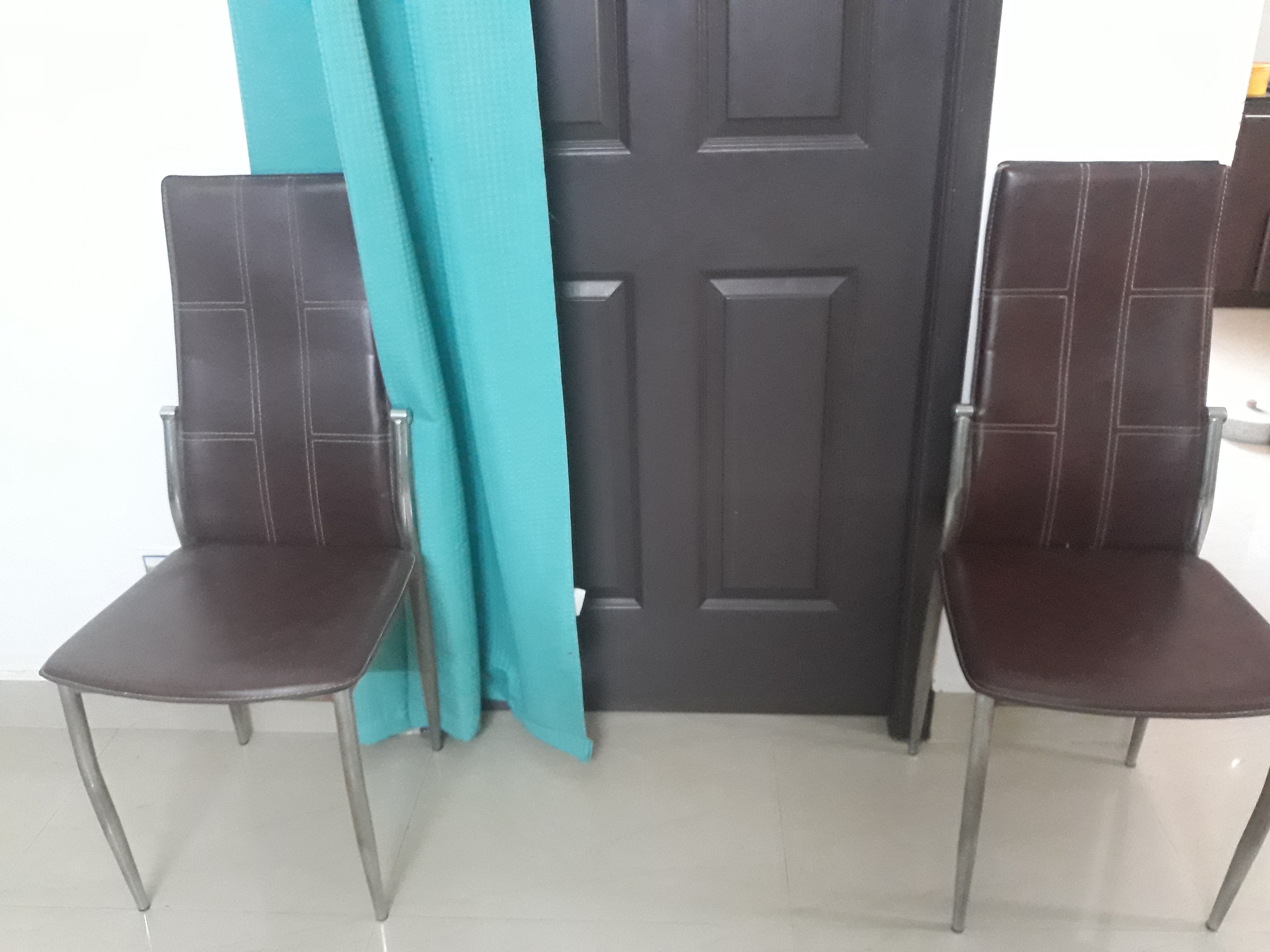}
 & can you find the chair & 1 chair on your left, 1 chair on your right \\
\noalign{\smallskip}\hline\noalign{\smallskip}

\centering
\includegraphics[width=2.5cm, height=2cm]{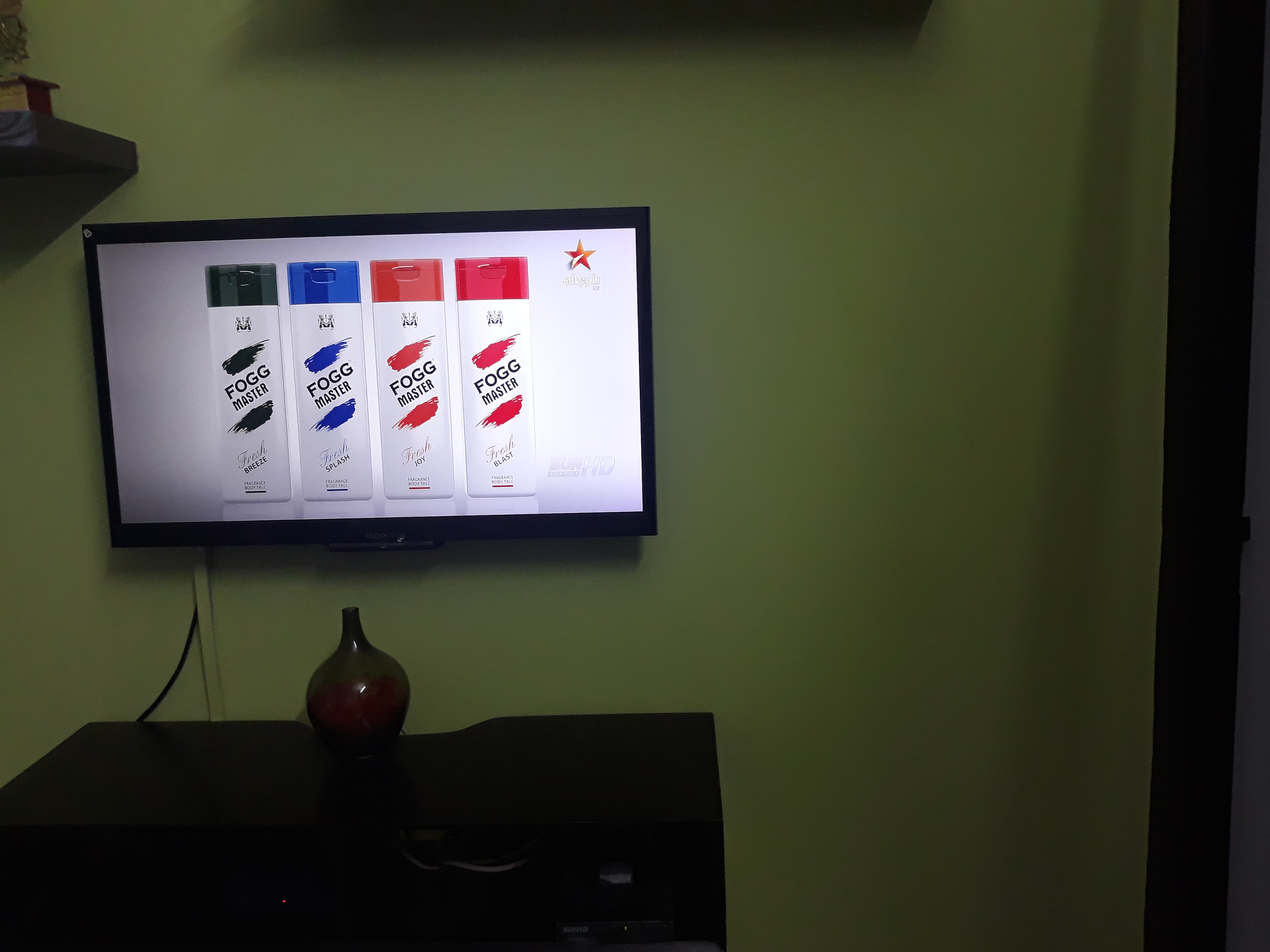}
 & can you find my chair & chair not found \\
\noalign{\smallskip}\hline\noalign{\smallskip}

\centering
\includegraphics[width=2.5cm, height=2cm]{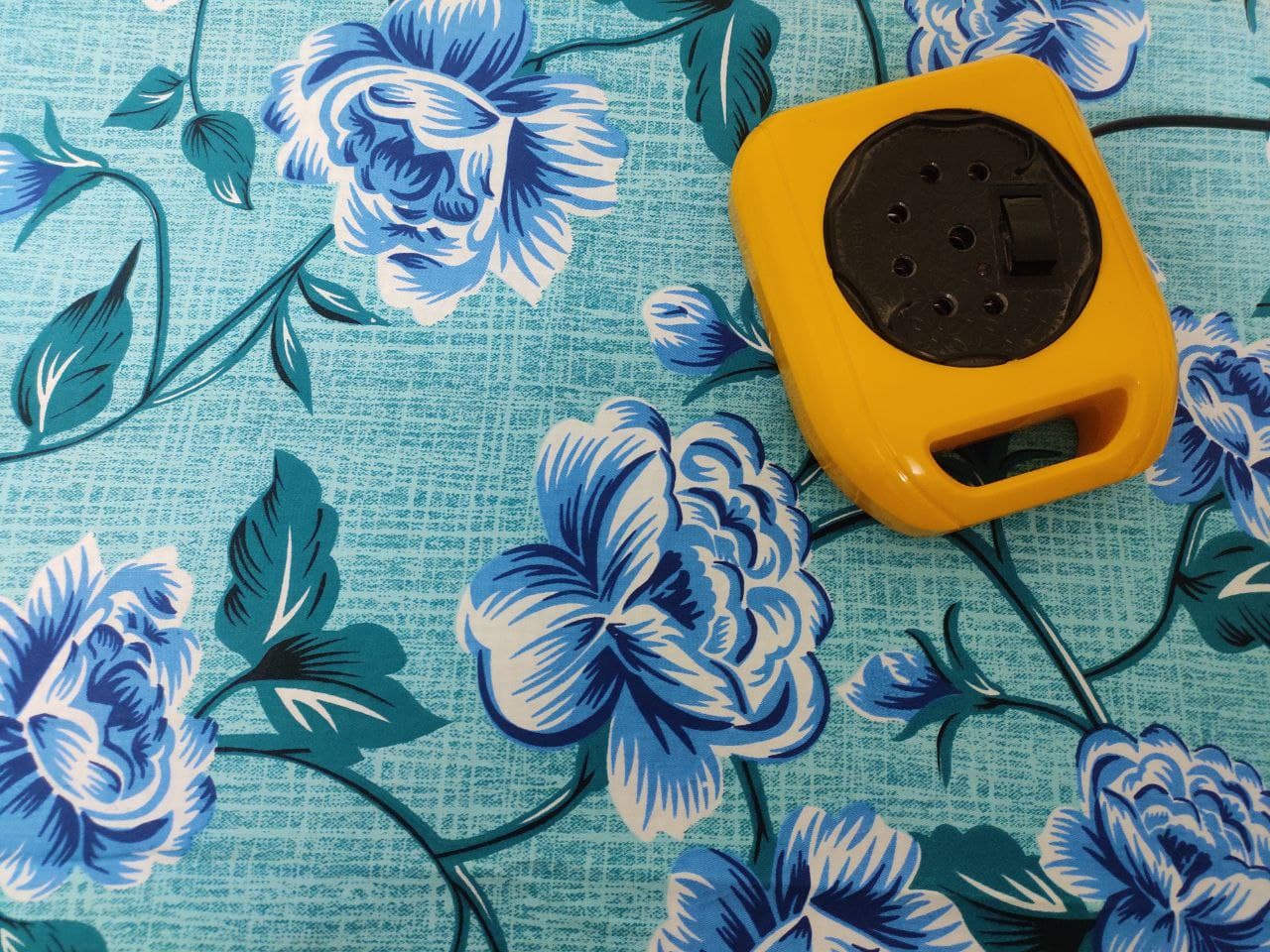}
 & can you find the extension box & object does not exist in the available classes \\
\noalign{\smallskip}\hline

\end{tabular}
\end{table}

\endgroup

\subsection{Reading Assistant}
The performance of Tesseract OCR Engine and Google Cloud Vision API for the Reading assistant task has been analyzed. Cosine similarity and BLEU score were used to evaluate the module's performance, and the results are tabulated in Table-4. 
\\Cosine similarity has been used to measure the correctness of the words \emph{(i.e)} whether the words detected by the algorithm are actually present in the image. It is not affected by order of the tokens. However, the order of the tokens is essential to convey meaning. Hence, to measure the semantic meaning and coherence of the identified text, BLEU-1, BLEU-2, BLEU-3 are computed. Sample outputs of text detection in a standard single-column image and a vertical-two-column-text image have been shown in figure 3 and figure 4, respectively.
\\The performance of Google Cloud Vision API was better than Tesseract. It was further enhanced with Z-score and K-means clustering to achieve better performance. This novel algorithm performed well on images containing texts with different fonts, orientations, outliers and multiple columns. Thus, Google Cloud Vision API with the Z-score and K-means clustering algorithm has been chosen to implement this module. The average latency for this module was 2s.

\begin{SCfigure}
  \centering
  \includegraphics[height=4cm, width=4cm]{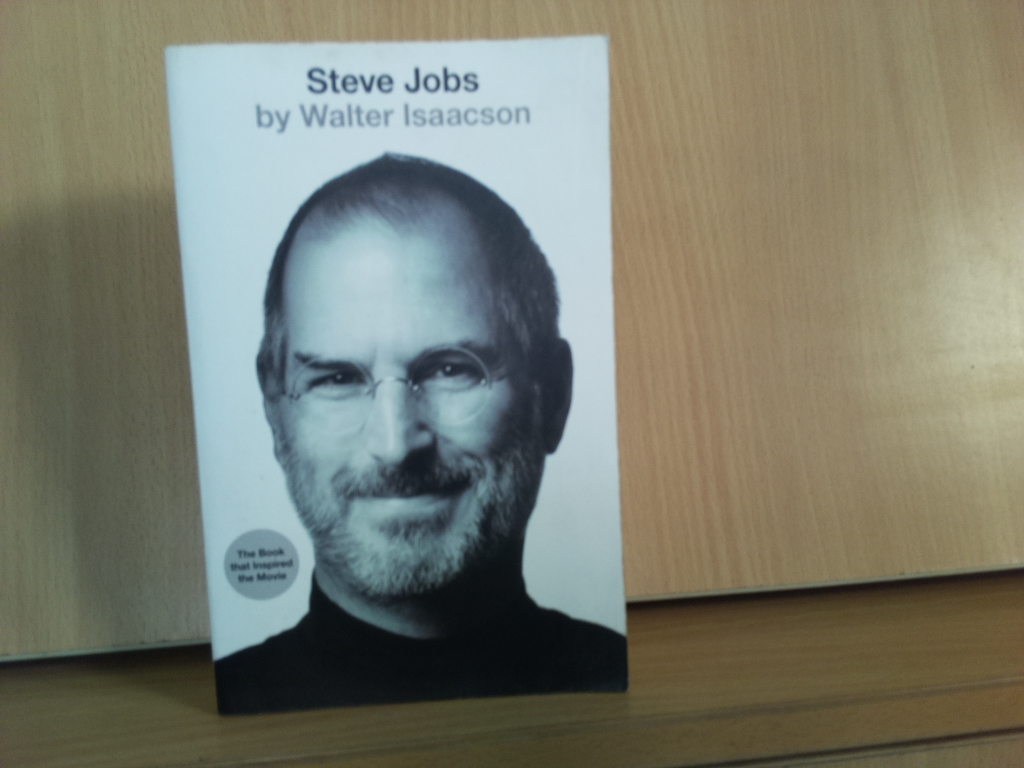}  
  \caption{Steve Jobs by Walter Isaacson \textbackslash n}
\end{SCfigure}

\begin{SCfigure}
  \centering
  \includegraphics[height=4cm, width=4cm]{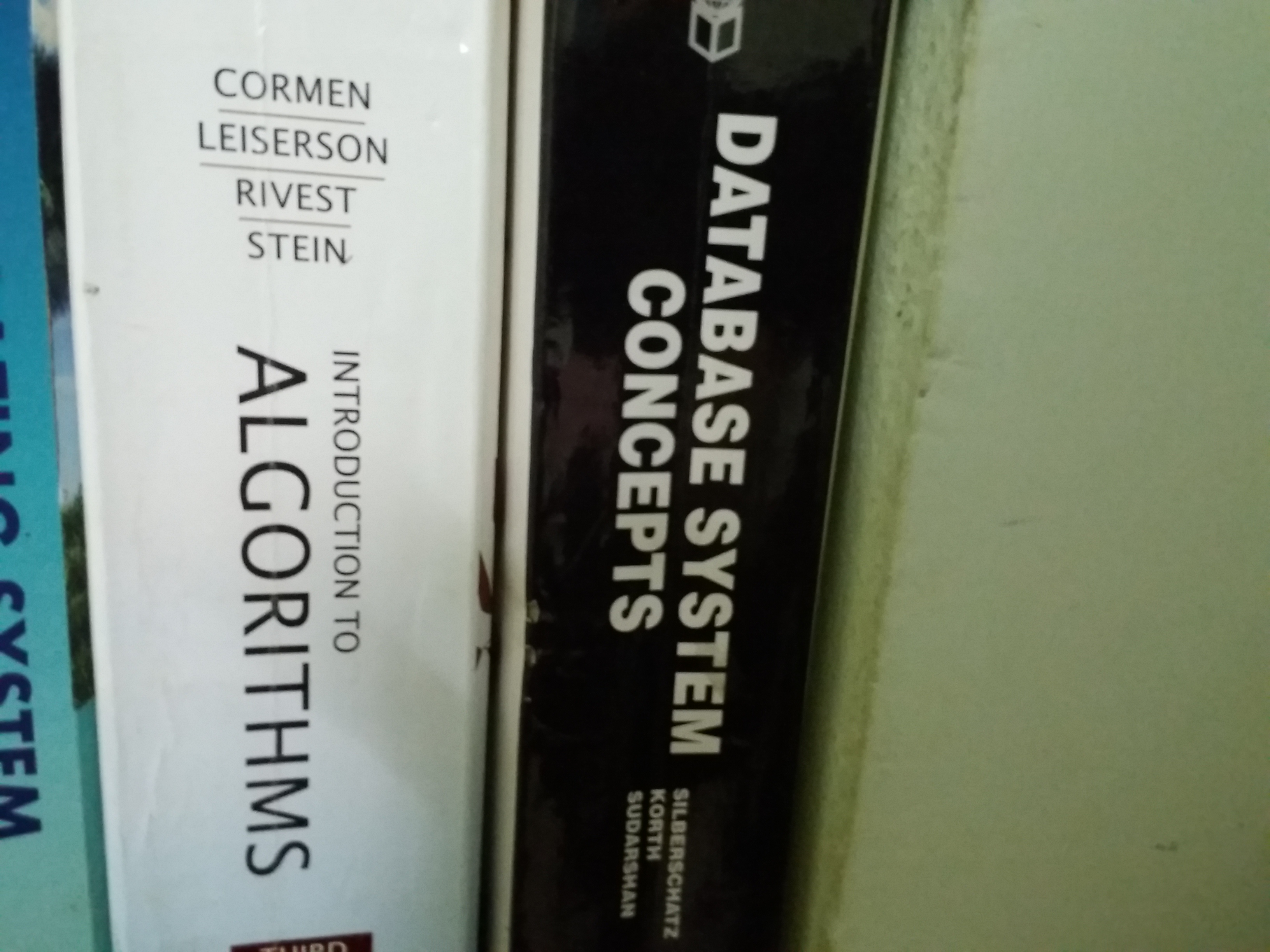}  
  \caption{CORMEN LEISERSON RIVEST STEIN \textbackslash n INTRODUCTION TO \textbackslash n ALGORITHMS \textbackslash n DATABASE SYSTEM CONCEPTS \textbackslash n SILBERSCHATZ KORTH SUDARSHAN \textbackslash n}
\end{SCfigure}

\begin{table}[h!]
\caption{Reading Assistant Metrics}
\label{tab:template}
\centering
\begin{tabular}{m{0.15\linewidth} m{0.13\linewidth} m{0.13\linewidth} m{0.2\linewidth}}
\noalign{\smallskip}\hline\noalign{\smallskip}

\centering
Category & Tesseract & Vision API & Vision API, Z-score, K-means clustering\\
\noalign{\smallskip}\hline\noalign{\smallskip}

\centering
Cosine Similarity & 0.24 & 0.73 & 0.81\\
\noalign{\smallskip}\hline\noalign{\smallskip}

\centering
BLEU-1 & 0.17 & 0.87 & 0.88\\
\noalign{\smallskip}\hline\noalign{\smallskip}

\centering
BLEU-2 & 0.12 & 0.69 & 0.72\\
\noalign{\smallskip}\hline\noalign{\smallskip}

\centering
BLEU-3 & 0.08 & 0.54 & 0.59\\
\noalign{\smallskip}\hline\noalign{\smallskip}

\centering
BLEU-4 & 0.06 & 0.40 & 0.48\\
\noalign{\smallskip}\hline

\end{tabular}
\end{table}


\subsection{Online Newspaper Reader}
The performance of the Newspaper Reader was measured using the Response time as the metric. It is the time interval between initiation of the module and reception of response. The mean response time for the online newspaper reading module has been approximately 26 seconds after a trial run of 10 times.

\section{Conclusion}

This paper proposes a smart wearable assistant for the visually challenged that is efficient, affordable and lightweight. No additional training or skill is required to operate it. It has been developed by leveraging the power of deep learning and cloud computing. VisBuddy has several features that make it versatile and enable it to assist the users with their daily activities. Each feature of the system has been explained in detail, with its performance metrics and use cases. Furthermore, the modules' performance with alternative deep learning models and OCR engines has also been discussed. Thus, the idea of integrating the concepts like image captioning, object detection, OCR and web scraping into a single system and deploying it on edge devices like the Raspberry Pi has been put forth in this paper. In conclusion, a prototype for an assistive device for the visually challenged has been discussed in this paper. This can be further enhanced by additional hardware and technological support to serve the visually challenged.

\section{Future Works}
The assistant presented in this paper can be enhanced to include other sophisticated features like GPS. The current object discovery module is limited to the identification of 80 objects. Including more daily life objects can improve the assistance in identifying more objects around the user. Finally, proximity sensors can also be added to measure and convey the distance between the user and the object.

\bibliographystyle{unsrt}  
\bibliography{references}

\begin{thebibliography}{10}

\bibitem{r1}
Muiz~Ahmed Khan, Pias Paul, Mahmudur Rashid, Mainul Hossain, and
  Md~Atiqur~Rahman Ahad.
\newblock An ai-based visual aid with integrated reading assistant for the
  completely blind.
\newblock {\em IEEE Transactions on Human-Machine Systems}, 50(6):507--517,
  2020.

\bibitem{r2}
Amey Hengle, Atharva Kulkarni, Nachiket Bavadekar, Niraj Kulkarni, and Rutuja
  Udyawar.
\newblock Smart cap: A deep learning and iot based assistant for the visually
  impaired.
\newblock pages 1109--1116, 2020.

\bibitem{r3}
Devashish~Pradeep Khairnar, Rushikesh~Balasaheb Karad, Apurva Kapse, Geetanjali
  Kale, and Prathamesh Jadhav.
\newblock Partha: A visually impaired assistance system.
\newblock In {\em 2020 3rd International Conference on Communication System,
  Computing and IT Applications (CSCITA)}, pages 32--37, 2020.

\bibitem{r9}
Roberto Manduchi.
\newblock Mobile vision as assistive technology for the blind: An experimental
  study.
\newblock In Klaus Miesenberger, Arthur Karshmer, Petr Penaz, and Wolfgang
  Zagler, editors, {\em Computers Helping People with Special Needs}, pages
  9--16, Berlin, Heidelberg, 2012. Springer Berlin Heidelberg.

\bibitem{r10}
S.~Kavya, Swathi, and Mimitha Shetty.
\newblock Assistance system for visually impaired using ai.
\newblock {\em International journal of engineering research and technology},
  7, 2019.

\bibitem{r11}
Kailas Patil, Qaidjohar Jawadwala, and Felix~Che Shu.
\newblock Design and construction of electronic aid for visually impaired
  people.
\newblock {\em IEEE Transactions on Human-Machine Systems}, 48:172--182, 2018.

\bibitem{r12}
Robert Katzschmann, Brandon Araki, and Daniela Rus.
\newblock Safe local navigation for visually impaired users with a
  time-of-flight and haptic feedback device.
\newblock {\em IEEE Transactions on Neural Systems and Rehabilitation
  Engineering}, PP:1--1, 01 2018.

\bibitem{r13}
Vidula Meshram, Kailas Patil, Vishal Meshram, and Felix Shu.
\newblock An astute assistive device for mobility and object recognition for
  visually impaired people.
\newblock {\em IEEE Transactions on Human-Machine Systems}, PP:1--12, 08 2019.

\bibitem{r4}
Kelvin Xu, Jimmy Ba, Ryan Kiros, Kyunghyun Cho, Aaron Courville, Ruslan
  Salakhutdinov, Richard Zemel, and Yoshua Bengio.
\newblock Show, attend and tell: Neural image caption generation with visual
  attention, 2016.

\bibitem{r17}
Karen Simonyan and Andrew Zisserman.
\newblock Very deep convolutional networks for large-scale image recognition,
  2015.

\bibitem{r18}
Mark Sandler, Andrew Howard, Menglong Zhu, Andrey Zhmoginov, and Liang-Chieh
  Chen.
\newblock Mobilenetv2: Inverted residuals and linear bottlenecks, 2019.

\bibitem{r8}
Christian Szegedy, Vincent Vanhoucke, Sergey Ioffe, Jonathon Shlens, and
  Zbigniew Wojna.
\newblock Rethinking the inception architecture for computer vision.
\newblock {\em CoRR}, abs/1512.00567, 2015.

\bibitem{r6}
Sepp Hochreiter and J\"{u}rgen Schmidhuber.
\newblock Long short-term memory.
\newblock {\em Neural Comput.}, 9(8):1735–1780, November 1997.

\bibitem{r15}
Junyoung Chung, {\c{C}}aglar G{\"{u}}l{\c{c}}ehre, KyungHyun Cho, and Yoshua
  Bengio.
\newblock Empirical evaluation of gated recurrent neural networks on sequence
  modeling.
\newblock {\em CoRR}, abs/1412.3555, 2014.

\bibitem{r5}
Tsung{-}Yi Lin, Priya Goyal, Ross~B. Girshick, Kaiming He, and Piotr
  Doll{\'{a}}r.
\newblock Focal loss for dense object detection.
\newblock {\em CoRR}, abs/1708.02002, 2017.

\bibitem{r16}
Andrew~G. Howard, Menglong Zhu, Bo~Chen, Dmitry Kalenichenko, Weijun Wang,
  Tobias Weyand, Marco Andreetto, and Hartwig Adam.
\newblock Mobilenets: Efficient convolutional neural networks for mobile vision
  applications, 2017.

\bibitem{r14}
Chirag Patel, Atul Patel, and Dharmendra Patel.
\newblock Optical character recognition by open source ocr tool tesseract: A
  case study.
\newblock {\em International Journal of Computer Applications}, 55:50--56, 10
  2012.

\bibitem{r7}
Kishore Papineni, Salim Roukos, Todd Ward, and Wei~Jing Zhu.
\newblock Bleu: a method for automatic evaluation of machine translation.
\newblock 10 2002.

\end{thebibliography}

\end{document}